\documentclass{article}

\usepackage[preprint]{corl_2026} % Uncomment for pre-prints (e.g., arxiv); This is like ``final'', but will remove the CORL footnote.

\usepackage{graphicx}
\usepackage{booktabs}
\usepackage{multirow}
\usepackage{colortbl}
\usepackage{arydshln}
\usepackage{wrapfig}
\usepackage{amsmath,amssymb,amsthm}
\usepackage{bm}
\usepackage{xcolor}
\usepackage{algorithm}
\usepackage{algorithmic}
\usepackage{placeins}
\usepackage{float}  % [H] for the appendix configuration tables
\usepackage{rotating}  % sidewaystable for the wide probe table in the appendix
\usepackage{enumitem}
\usepackage{hyperref} % If used, must come BEFORE cleveref
\usepackage{cleveref} % Always load last!

\newcommand{\lewam}{LeWAM}
\newcommand{\dscem}{DS-CEM}
\newcommand{\ds}{DS}  % diffusion steering; the instances are \dscem{}, DS-MPPI, DS-CMA-ES
\newcommand{\cmark}{\checkmark}  % diffusion steering: any sampling-based planner run over the policy head's noise
\newcommand{\genwam}{RecWAM}  % the generative (reconstruction-based) baseline: the Cosmos Policy architecture at LeWAM's size
\newcommand{\R}{\mathbb{R}}

\title{Diffusion-Steering-Based MPC with a JEPA World Action Model}

\author{
  Shashank Hegde\\
  \And
  Alexander Popov\\
  \And
  Elie Aljalbout\\
  \And
  Nikolai Smolyanskiy \\
  \AND
    NVIDIA
}

\begin{document}
\maketitle

%===============================================================================

\begin{abstract}
    % The purpose of this document is to provide both the basic paper template and submission guidelines. Abstracts should be a single paragraph, between 4--6 sentences long, ideally. Gross violations will trigger corrections at the camera-ready phase.
    % World action models (WAMs) predict actions and future observations, typically from a reconstruction-based representation that carries noisy, redundant information which can complicate downstream predictions.
% We introduce \lewam{}, a bidirectional transformer for forward, backward, inverse dynamics and policy prediction, on a decoder-free JEPA latent trained end-to-end through all four modes. 
% We see the following benefits:
% 1) Alignment: it produces a latent state which better focuses on the robot and object states while successfully ignoring irrelevant parts of the scene (distractors) compared to a regular Le World Model (a forward-only JEPA) world model.
% 2) Acting: Closed-loop evaluations of LeWAM match or exceed policies produced by JEPA and WAM style models.
% 3) Planning: \lewam{} enables planning in the noise space, increasing overall closed-loop performance compared to classical MPC based WAMs.
World action models (WAMs) predict actions and future observations, typically from a reconstruction-based representation that carries noisy, redundant information which can complicate downstream predictions.
We introduce \lewam{}, a bidirectional transformer for forward, backward, inverse dynamics and policy prediction, on a decoder-free JEPA latent trained end-to-end through all four modes.
We see the following benefits:
1) Alignment: linear probes read robot and object state from \lewam{}'s latent better than from a regular Le World Model (a forward-only JEPA world model), while the latent ignores visual distractors as well as LeWM does and far better than a reconstruction-based WAM.
2) Acting: Closed-loop evaluations of \lewam{} match a regular flow-matching policy trained on the same encoder at matched size, while also providing a world model.
3) Planning: Sampling raw actions when planning with WAMs lets MPC exploit dynamics-model inaccuracies; planning in the noise space of the policy head instead improves the closed-loop performance of these WAMs.
\end{abstract}


%===============================================================================

% \section{Introduction}
	
%     Submission to CoRL 2026 will be entirely electronic, via a web site (not email). Information about the submission process and \LaTeX{} templates are available on the conference web site at \url{https://corl.org/}. For camera ready submission, use the \texttt{final} option for the \texttt{\textbackslash usepackage} command. 

\section{Introduction}
% World Action Models (WAMs) jointly predict future states and the actions required to reach them~\citep{zhu2025uwm,nvidia2025cosmospolicy}. Recent generative WAMs, such as Unified World Models~\citep{zhu2025uwm} and Cosmos Policy~\citep{nvidia2025cosmospolicy}, shape their latent representations via pixel reconstruction. Consequently, the latent space must encode every visual detail the decoder will render, whether or not it is needed for the task. In contrast, the joint-embedding predictive architecture (JEPA)~\citep{lecun2022path,assran2023ijepa,bardes2024vjepa} models dynamics in latent space alone.

% Inspired by Le World Model (LeWM)~\citep{lewm2026}, we propose \lewam{}, a unified, reconstruction-free WAM. Its predictor is conditioned on actions, while its encoder is trained end-to-end from the task's own pixels under a regularizer that prevents representational collapse. By evaluating on human-generated PushT and robomimic data with visual distractors, we show that: (i) \lewam{}'s latent space is as dynamically informative as that of reconstruction-based WAMs, while remaining as clear of visual distractors as LeWM's; (ii) its policy matches or exceeds that of reconstruction-based WAMs and matches regular behavioral cloning (BC) policies, while additionally providing a world model for dynamics prediction; and (iii) its action head lets us run MPC with the learned dynamics model by sampling in the policy's noise space instead of action space, giving a performance increase at inference time.

Robotic control increasingly relies on predictive models to anticipate environmental dynamics. Traditional world models learn a latent space through image reconstruction, subsequently training policies conditioned on these representations \citep{hafner2020dreamer, hafner2023dreamerv3}. 
A recent evolution of this paradigm, termed world action models (WAM), explicitly couples future-state prediction with action generation to ground control directly in learned dynamics \citep{zhu2025uwm, nvidia2025cosmospolicy, ye2026dreamzero}. While these generative approaches successfully integrate vision and action, they predominantly rely on pixel-level reconstruction to shape their underlying representations \citep{zhu2025uwm}. Consequently, their latent spaces are forced to encode every visual detail a decoder might render, regardless of its relevance to the task \citep{zhu2025uwm}. In visually complex environments, this architectural choice inherently wastes model capacity on noisy, task-irrelevant information, complicating downstream predictive tasks \citep{zhang2018natural, stone2021distracting}.

To circumvent the limitations of reconstruction, alternative frameworks such as the joint-embedding predictive architecture model dynamics exclusively in latent space \citep{lecun2022path, assran2023ijepa, bardes2024vjepa}. By discarding the pixel decoder entirely, these architectures inherently resist task-irrelevant appearance details \citep{deng2022dreamerpro, zhang2021dbc}. Recent instantiations demonstrate stable, end-to-end training from pixels using regularization techniques that prevent representational collapse without relying on heuristics \citep{lewm2026, balestriero2025lejepa}. Despite these representational advantages, existing frameworks in this category typically separate the dynamics model from the policy, lacking the unified multi-modal structure found in modern WAMs \citep{lewm2026}.

Bridging this gap, we propose LeWAM, a unified, reconstruction-free world action model. Building upon foundational decoder-free predictors, LeWAM introduces a bidirectional transformer capable of simultaneous forward dynamics, backward dynamics, inverse dynamics, and policy prediction \citep{lewm2026}. The predictor is conditioned on actions, while the encoder is trained end-to-end directly from task pixels under a strict regularizer. Because the architecture eschews pixel decoders, tokenizers, and pixel-space losses, its representations are shaped entirely by the dual objectives of accurate dynamics prediction and robust action generation \citep{lipman2023flow, liu2023rectified}.

Integrating a generative policy within a world model naturally invites test-time intervention through planning \citep{janner2022diffuser, yang2024diffusiones}. However, executing sampling-based model predictive control over raw actions often fails in practice \citep{chua2018pets}. When optimizing over action space, the planner tends to exploit inevitable inaccuracies within the learned dynamics model, leading to plans that optimize for hallucinated futures and subsequently fail during deployment \citep{argenson2021mbop}. To resolve this, we shift the decision variable of the sampling-based planner from the action space to the deterministic noise space of the flow-matching policy \citep{wagenmaker2025dsrl, simsir2026lpds, patil2026goldenticket}. This diffusion steering approach searches strictly within the world model's imagination, evaluating candidates decoded at imagined states \citep{hansen2024tdmpc2}. By projecting proposals onto the sampler's typical set, the search remains constrained to actions the policy head was explicitly trained to produce.

We evaluate our approach on human-generated demonstration datasets augmented with visual distractors \cite{chi2023diffusion, robomimic2021}. Our investigation yields the following primary contributions:
\begin{itemize}
    \item \textbf{Alignment:} We demonstrate representational alignment; the latent space captures dynamic robot and object states as effectively as reconstruction-based baselines, while successfully ignoring visual distractors.
    \item \textbf{Acting:} We validate robust acting capabilities; the integrated flow-matching policy matches the performance of standard behavioral cloning policies while simultaneously providing a world model for dynamics prediction.
    \item \textbf{Planning:} We establish a stable planning paradigm; steering the generative process by sampling within the policy's noise space prevents the exploitation of dynamics errors and significantly improves closed-loop performance during inference \citep{rubinstein1999cem, williams2017mppi, hansen2016cmaes}.
\end{itemize}
\section{Related work}
\label{sec:related}

% \Cref{tab:rw-wm} places \lewam{} among the world models and world action models.
\begin{table}[t]
\caption{World models and world action models as published: what shapes the latent and what one network does. LeWM has no policy head. Dreamer \citep{hafner2023dreamerv3}; TD-MPC2 \citep{hansen2024tdmpc2}; DINO-WM and V-JEPA 2-AC \citep{zhou2024dinowm,assran2025vjepa2}; LeWM \citep{lewm2026}; UWM \citep{zhu2025uwm}; Cosmos Policy \citep{nvidia2025cosmospolicy}. 
% ``Action loss to encoder'': a policy or inverse-dynamics loss on the action itself whose gradient reaches the encoder (a value loss, as in TD-MPC2, does not count). ``One backbone'': one network serving every mode (forward, inverse, policy) by what is hidden. \cmark: present; --: absent.
}
\label{tab:rw-wm}
\centering\footnotesize
\setlength{\tabcolsep}{2.5pt}
\begin{tabular}{@{}lcccccccc@{}}
\toprule
 & decoder- & action loss & policy & forward & inverse & backward & multimodal & one \\
 & free & to encoder & head & dyn. & dyn. & dyn. & head & backbone \\
\midrule
Dreamer & -- & -- & \cmark & \cmark & -- & -- & -- & -- \\
TD-MPC2 & \cmark & -- & \cmark & \cmark & -- & -- & -- & -- \\
DINO-WM / V-JEPA 2-AC & \cmark & -- & -- & \cmark & -- & -- & -- & -- \\
LeWM & \cmark & -- & -- & \cmark & -- & -- & -- & -- \\
UWM & -- & -- & \cmark & \cmark & \cmark & -- & \cmark & \cmark \\
Cosmos Policy & -- & -- & \cmark & \cmark & \cmark & -- & \cmark & \cmark \\
\lewam{} (ours) & \cmark & \cmark & \cmark & \cmark & \cmark & \cmark & \cmark & \cmark \\
\bottomrule
\end{tabular}
\end{table}

\paragraph{World models and planning in latent space.} World Models such as Dreamer~\citep{ha2018worldmodels,hafner2020dreamer,hafner2023dreamerv3} learn a recurrent latent space primarily via reconstruction and then train policies conditioned on latents in this space. 
The main shortcoming of reconstruction-based world modeling is in the capacity wasted on noisy and task-irrelevant information. 
Many approaches have been proposed to address this problem.
For instance, DreamerPro and bisimulation methods \citep{deng2022dreamerpro,zhang2021dbc} drop the decoder to resist task-irrelevant appearance \citep{zhang2018natural,stone2021distracting}.
TD-MPC2~\citep{hansen2022tdmpc,hansen2024tdmpc2} plans by MPPI over a decoder-free latent and value.
JEPA world models~\citep{zhou2024dinowm,bardes2024vjepa,assran2025vjepa2} plan by CEM-MPC on a goal-latent distance, and LeWM~\cite{lewm2026} trains its encoder end-to-end under SIGReg~\citep{balestriero2025lejepa}. \lewam{} adds the action head and multi-mode skeleton to LeWM's latent. We present a clear comparison to these methods in \Cref{tab:rw-wm}. Note that the head on LeWM's frozen latent we present in our experiments is our control, not an intrinsic part of LeWM.

\iffalse % Table of test-time planners, hidden: the comparison is made in the text below
\begin{table}[t]
\caption{Test-time planners: where the search runs and what it needs. CEM and MPPI over actions \citep{lewm2026,zhou2024dinowm}; policy-prior MPC \citep{hansen2024tdmpc2,argenson2021mbop,hamel2026flowmpc}; best-of-$N$ ranking \citep{nvidia2025cosmospolicy,nakamoto2024vgps,zhou2024dmpc}; DSRL \citep{wagenmaker2025dsrl}; Golden Ticket \citep{patil2026goldenticket}; Diffusion-ES \citep{yang2024diffusiones}. \cmark: yes; --: no.}
% \vspace{-2mm}
\label{tab:rw-plan}
\centering\small
\setlength{\tabcolsep}{4pt}
\begin{tabular}{@{}lcccccc@{}}
\toprule
 & policy & noise as & online & scored in & goal only, & any \\
 & samples & variable & & imagination & no value & update \\
\midrule
CEM / MPPI over actions & -- & -- & \cmark & \cmark & \cmark & \cmark \\
policy-prior MPC & -- & -- & \cmark & \cmark & -- & -- \\
best-of-$N$ ranking & \cmark & -- & \cmark & \cmark & -- & -- \\
DSRL & \cmark & \cmark & \cmark & -- & -- & -- \\
Golden Ticket & \cmark & \cmark & -- & -- & -- & -- \\
Diffusion-ES & \cmark & \cmark & \cmark & -- & -- & -- \\
\ds{} (ours) & \cmark & \cmark & \cmark & \cmark & \cmark & \cmark \\
\bottomrule
\end{tabular}
\end{table}
\fi
\paragraph{World action models.} 
One recent trend in robotics is to couple future-state prediction with action generation, grounding robot control in learned dynamics in an architecture referred to as world action models. 
Dreamitate adapts a video generator to task-specific human demonstrations and tracks tools in its predictions to obtain executable trajectories \citep{liang2024dreamitate}. 
Unified World Models~(UWM) makes this connection explicit through independently controlled video and action diffusion, allowing one model to serve as a policy, forward model, or inverse model~\citep{zhu2025uwm}. 
Other methods such as DreamZero exploit pretrained video priors to generalize to unseen tasks and motions \citep{ye2026dreamzero}. 
How such priors guide action varies, for instance, Cosmos Policy incorporates actions, future states, and values into the video model's native diffusion process, enabling both control and planning \citep{nvidia2025cosmospolicy}. 

\paragraph{Steering a generative policy.} 

Frozen diffusion and flow-matching policies \citep{chi2023diffusion,lipman2023flow,liu2023rectified,black2024pi0} offer several points of intervention for test-time control. 
Guidance steers the sampling process itself~\citep{janner2022diffuser}, whereas candidate ranking selects among policy samples using a learned value~\citep{nakamoto2024vgps,nvidia2025cosmospolicy} or world-model rollouts~\citep{zhou2024dmpc,qi2025gpc}. 
Rather than merely selecting a proposal, MPC can refine it through MPPI in \emph{action} space, potentially leaving the policy's support~\citep{argenson2021mbop,hamel2026flowmpc}, while Diffusion-ES alternates simulator-based scoring with re-noising and denoising~\citep{yang2024diffusiones}. 
A different route is to optimize the noise that generates the action. 
DSRL learns a policy over this noise through environment interaction, without a world model or planning~\citep{wagenmaker2025dsrl,simsir2026lpds}. 
Golden Ticket instead searches offline with CEM, using real task-reward rollouts to find one task-specific noise vector that is reused at every step \citep{patil2026goldenticket}.%; related image-generation methods also optimize initial noise \citep{guo2024initno,schweiger2026trs}. 
~\ds{} builds on this formulation rather than introducing a new optimizer: both its decision variable and \dscem{}'s update match Golden Ticket's. 
The distinction lies in when and where search happens, and how candidates are evaluated. 
At every decision, \ds{} searches within world-model imagination, decoding action chunks at imagined states and scoring them by goal-latent distance, without search-time environment interaction or a learned value function. 
This formulation accommodates different sampling-based updates such as CEM~\citep{rubinstein1999cem,chua2018pets}, MPPI~\citep{williams2017mppi}, and CMA-ES~\citep{hansen2016cmaes}.

% \vspace{-3mm}
\section{\lewam{}}
\label{sec:method}
\begin{figure}[t]
\centering
\includegraphics[width=\linewidth]{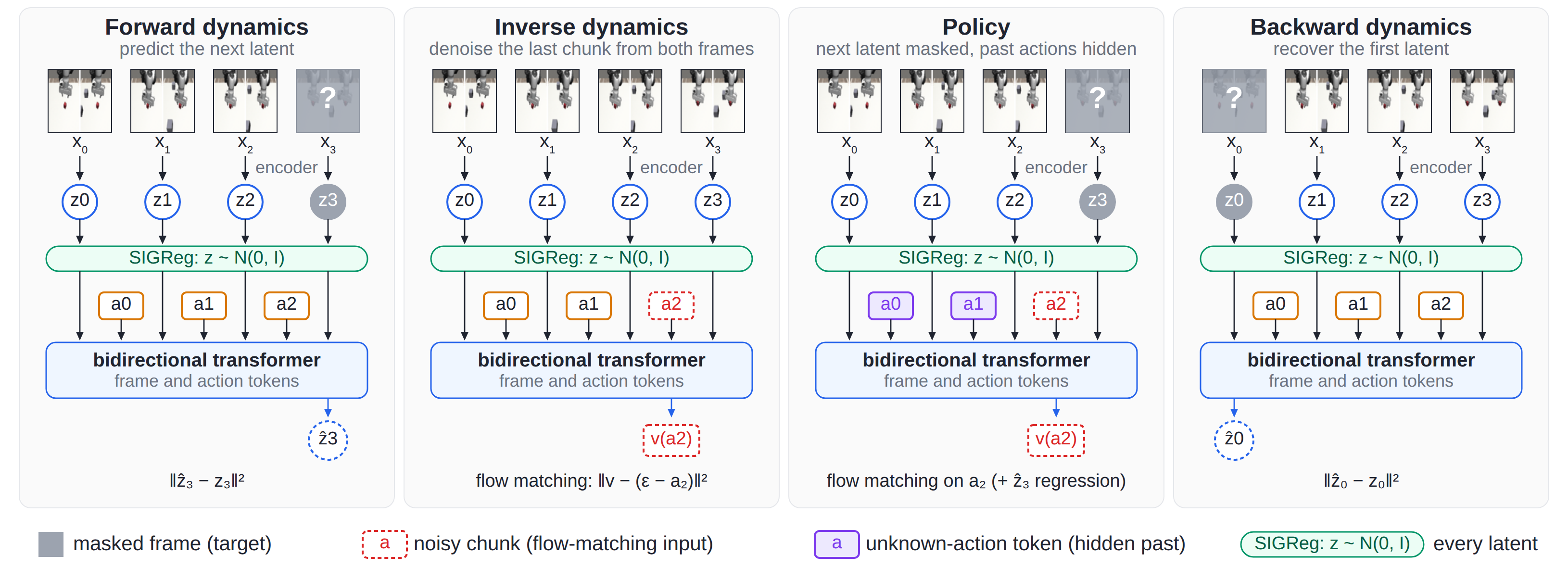}
% \vspace{-6mm}
\caption{The four training modes on one window: the mode is set by what is masked at the transformer's input. Nothing is decoded to pixels.}
% \vspace{-4mm}
\label{fig:modes}
\end{figure}

We use a multi-mode skeleton following UWM and Cosmos Policy. This model operates like the predictor in LeWM (\Cref{fig:modes}). 
% See \Cref{sec:arch} for more details about the architecture.

% \vspace{-2mm}

\paragraph{Inputs and notation.} A training window contains four frames, $o_{t-2:t+1}$, at a frameskip of five, and the three 5-step action chunks between them, $a_{t-2:t} \in \mathbb{R}^{A}$. The chunk dimension $A$ is five steps times the per-step action dimension ($A=10$ on PushT, $35$ on robomimic, $70$ on Transport), z-scored per dimension. The first three frames form the context, with latents $\mathbf{z}_t = (z_{t-2}, z_{t-1}, z_t)$; the fourth frame is the prediction target, and the final chunk $a_t$ is the chunk the policy must produce.

\paragraph{Modes.} \lewam{} masks specific inputs to serve four training modes: (i) \emph{Forward}: the final frame's latent is replaced by a mask token, all actions are clean, and the backbone regresses $z_{t+1}$ via MSE against the live encoder's output; (ii) \emph{Backward}: masks the first frame's latent, regressing it from the later frames and all actions; (iii) \emph{Inverse}: keeps all frames clean and replaces the target chunk $a_t$ with a noisy flow-matching sample whose velocity the backbone predicts, inferring the action that bridges past and future states; and (iv) \emph{Policy}: masks the final frame and replaces $a_t$ with a noisy flow-matching sample, so both losses apply and the model must deduce the action from the frames while predicting the future latent. In this mode the past action tokens ($a_{t-2:t-1}$) are replaced by a learned unknown-action token, so the policy cannot copy the demonstrator's recent motion.

During training, every mode is executed on every batch: the encoder processes the inputs once, the backbone runs independently for each mode, and the losses are summed. The Forward and Backward modes, alongside SIGReg~\citep{balestriero2025lejepa}, impose a JEPA structure on the latent space, while the Inverse and Policy modes make the space more task-aligned.

\paragraph{Action Generation.}
Action generation employs rectified flow~\citep{liu2023rectified}. During training, we sample $\sigma \sim U(0,1)$ and construct a noisy action $a_\sigma = (1-\sigma)a + \sigma\epsilon$, training the backbone to predict the velocity target $\epsilon - a$. At test time, the action chunk is integrated from standard noise $\epsilon \sim \mathcal{N}(0, I)$ using eight Euler steps. This integration forms a deterministic mapping, denoted $a = g(\epsilon; \mathbf{z}_t)$.

Notably, unlike reconstruction-based WAMs, our architecture operates without a pixel decoder, tokenizer, or any pixel- or token-space loss. As a result, the latent representations are shaped entirely by the dual objectives of dynamics prediction and action generation.

% \subsection{Architecture}
% \label{sec:arch}

\paragraph{Encoder and latent.} Each frame is independently encoded at a resolution of $224$ pixels using a ViT-tiny/14. 
We use the DINOv2 small architecture \citep{oquab2024dinov2}, but train it entirely from scratch without pretrained weights. 
The encoder's [CLS] token passes through a BatchNorm MLP projector to produce a single latent vector, $z_t \in \R^{384}$. 
This vector serves as the model's sole representation: it is read by the probes, compared by the planner, and regressed by the dynamics model.
Similar to LeWM, this latent is taken directly from the live encoder output, without relying on an exponential moving average (EMA) target network or a stop-gradient operation. 
To prevent representation collapse, we apply SIGReg \citep{balestriero2025lejepa}, which regularizes the model by pushing the batch of latents toward an isotropic Gaussian distribution.

\begin{figure}[h]
\centering
\includegraphics[width=\linewidth]{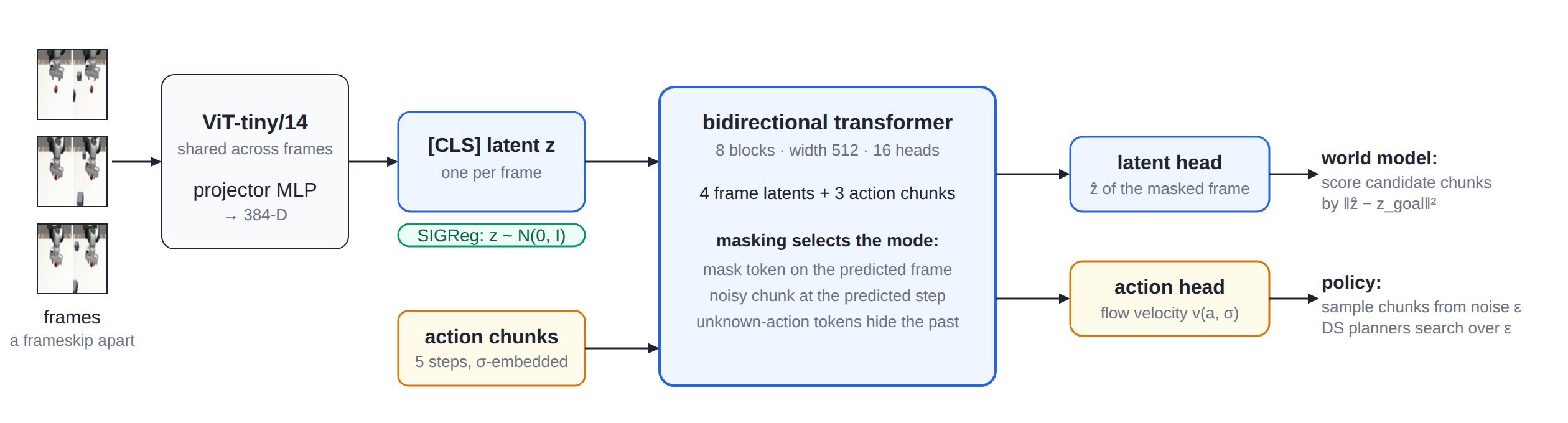}
\caption{\lewam{}: a shared ViT-tiny encoder with SIGReg on its single latent per frame, and one bidirectional transformer that serves the four training modes by masking. No pixel decodeding.}
\label{fig:arch}
\end{figure}

\paragraph{Backbone} A bidirectional transformer of depth 8, width 512 and 16 heads reads the three frame latents and three action tokens with modality, frame and $\sigma$ embeddings, and serves four modes by masking. 
\Cref{fig:arch} shows the \lewam{} architecture of \Cref{sec:method}: a shared ViT-tiny encoder with SIGReg on its single latent per frame, and one bidirectional transformer that serves the four training modes of \Cref{fig:modes} by masking.
% \vspace{-2mm}
\section{Diffusion steering: planning in the policy's noise space}
\label{sec:dscem}

% \vspace{-2mm}
\subsection{Goal-conditioned planning with a latent distance}
Given context $\mathbf{z}_t$ and a goal latent $z_g$ retrieved from a demonstration, following \cite{zhou2024dinowm,lewm2026}, we score a chunk $a$ by
\begin{equation}
\hat c(a) \;=\; \big\lVert \hat f(\mathbf{z}_t, a) - z_g \big\rVert_2 ,
\label{eq:cost}
\end{equation}
with $\hat f$ the forward mode, and minimize it over $H$ chunks by a sampling-based planner.
The latter proposes $S$ plans, scores each on its rolled-out final latent, updates the distribution, repeats this process $I$ times, and executes the chosen plan's first chunk before replanning. 
The update rule defines the planner.
% For instance, CEM refits a Gaussian to the $K$ elites and executes the best, MPPI softmax-weights candidates by cost and executes their mean, and CMA-ES adapts a full covariance and step size \citep{rubinstein1999cem,williams2017mppi,hansen2016cmaes}. 
Sampling-based MPCs originally search over raw actions $a_{1:H} \in \R^{AH}$. 

% \vspace{-2mm}
% \subsection{Motivation: Failure of Action-Space MPC with LeWM}
% \label{sec:break}
% While action-space sampling-based MPC has been shown to work effectively with LeWM on large, scripted benchmarks, it fails when applied to complex human demonstration data. We hypothesize that human behavior forms a thin, multimodal manifold within the raw action space. Standard Gaussian sampling used by action-space MPC rarely lands on this manifold, forcing LeWM to evaluate out-of-distribution actions. Because the latent goal-distance metric inadvertently rewards unconstrained, extrapolated moves, the planner rapidly drifts into the world model's extrapolation errors—optimizing for hallucinated futures that fail during deployment.
\subsection{Motivation: Failure of Action-Space MPC with LeWM}
\label{sec:break}
% \vspace{-2mmk}

While action-space sampling-based MPC has been shown to work with LeWM, it fails on the smaller human-sourced datasets studied here (see \Cref{fig:avg_cle}). We hypothesize that the dynamics model is imperfect with less data, and compounds errors during planning rollouts. Gaussian action-sampling MPC on such a model optimizes for hallucinated futures, and the planned actions fail during deployment.
\subsection{Diffusion steering}
\label{sec:ds}
% \vspace{-2mm}

Unlike regular world models, a flow-matching WAM has a policy head whose sampler is a deterministic map $g(\cdot; \mathbf{z})$ from a Gaussian draw $\epsilon \in \R^{A}$, to a chunk. 
Diffusion steering (DS) moves the decision variable of any sampling-based planner from the actions to that noise (any diffusion or flow sampler provides the map). 
The plan is denoted by $\epsilon_{1:H}$. Chunk $h$ is decoded at the state imagined after chunks $1..h{-}1$,
\begin{equation}
a_h = g(\epsilon_h;\, \hat{\mathbf{z}}_{h-1}), \qquad \hat z_h = \hat f(\hat{\mathbf{z}}_{h-1}, a_h), \qquad \hat{\mathbf{z}}_h = \mathrm{shift}(\hat{\mathbf{z}}_{h-1}, \hat z_h), \qquad \hat{\mathbf{z}}_0 = \mathbf{z}_t ,\label{eq:rollout}
\end{equation}
where $\mathrm{shift}$ drops the oldest context latent and appends the predicted one, and the plan is scored by \eqref{eq:cost} on its final latent $\hat z_H$. The search starts at the policy's prior $\mathcal{N}(0, I)$, so the first iteration of every \ds{} planner is best-of-$N$ proposal ranking with $N = S$ over projected noise \citep{nvidia2025cosmospolicy,nakamoto2024vgps}. \dscem{}, DS-MPPI and DS-CMA-ES share one budget ($64 \times 3$, horizon 5), the only difference between them is in the update rule.

% \vspace{-2mm}
\subsection{Projection onto the sampler's typical set}
\label{sec:variants}
% \vspace{-2mm}

The head was trained to start from draws of $\mathcal{N}(0, I)$, which concentrate on the sphere of radius $\sqrt{A}$; as a planner narrows its search its proposals drift toward a mean of much smaller norm, which for a rectified-flow head acts like sampling at reduced temperature (mode-seeking). 
\emph{Projected} \ds{} maps every proposal back onto the sphere, $\epsilon \leftarrow \sqrt{A}\,\epsilon/\lVert\epsilon\rVert$, before decoding. 
As a consequence, the search is over \emph{directions}, every decoded candidate is one the head was trained to produce, and the temperature effect is given up. 
The search distribution stays Euclidean; only proposals are projected. 
The search distribution stays Euclidean; only proposals are projected (ablated in \Cref{tab:projection}).

Putting it altogether, we get \Cref{alg:dscem}.  
\begin{algorithm}[t]
\caption{Diffusion steering of a sampling-based planner $\mathcal{P}$ (one decision)}
\label{alg:dscem}
\begin{algorithmic}
\STATE {\bfseries Input:} context $\mathbf{z}_t$, goal $z_g$, horizon $H$, samples $S$, iterations $I$, policy map $g$, forward mode $\hat f$, planner $\mathcal{P}$ with \textsc{propose}, \textsc{update}, \textsc{select} (CEM: Gaussian refit to $K$ elites, best elite; MPPI: softmax reweighting, weighted mean; CMA-ES: covariance and step-size adaptation, best sample)
\STATE $\theta \leftarrow$ the policy's prior $\mathcal{N}(0, I)$ over $\epsilon_{1:H}$
\FOR{$i = 1$ {\bfseries to} $I$}
  \STATE $\epsilon^{(1:S)}_{1:H} \leftarrow \textsc{propose}(\theta)$
  \STATE $\epsilon^{(s)}_h \leftarrow \sqrt{A}\,\epsilon^{(s)}_h / \lVert \epsilon^{(s)}_h \rVert$ for every $s, h$ \hfill (projection onto the sampler's typical set)
  \FOR{each $s$}
     \STATE $\hat{\mathbf{z}} \leftarrow \mathbf{z}_t$
     \FOR{$h = 1$ {\bfseries to} $H$}
        \STATE $a^{(s)}_h \leftarrow g(\epsilon^{(s)}_h; \hat{\mathbf{z}})$; \quad $\hat z \leftarrow \hat f(\hat{\mathbf{z}}, a^{(s)}_h)$; \quad $\hat{\mathbf{z}} \leftarrow \mathrm{shift}(\hat{\mathbf{z}}, \hat z)$
     \ENDFOR
     \STATE $c^{(s)} \leftarrow \lVert \hat z - z_g \rVert$
  \ENDFOR
  \STATE $\theta \leftarrow \textsc{update}(\theta, \epsilon^{(1:S)}, c^{(1:S)})$
\ENDFOR
\STATE {\bfseries return} the first chunk of $\textsc{select}(\theta, \epsilon^{(1:S)}, c^{(1:S)})$, decoded at $\mathbf{z}_t$
\end{algorithmic}

\end{algorithm}

\begin{figure}[t]
\centering\small
\setlength{\tabcolsep}{2pt}
\vspace{6mm}

\begin{tabular}{cccccc}
\includegraphics[width=0.155\linewidth]{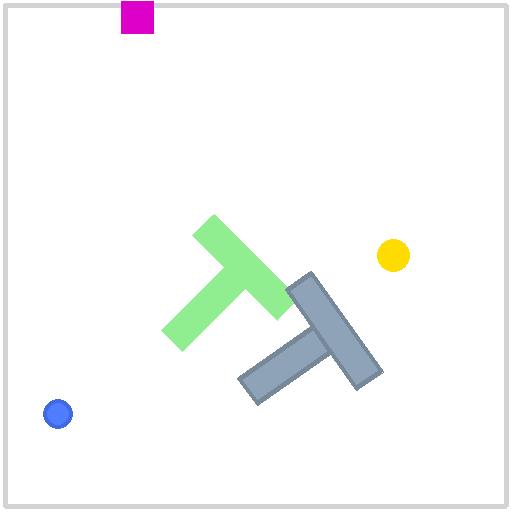} & \includegraphics[width=0.155\linewidth]{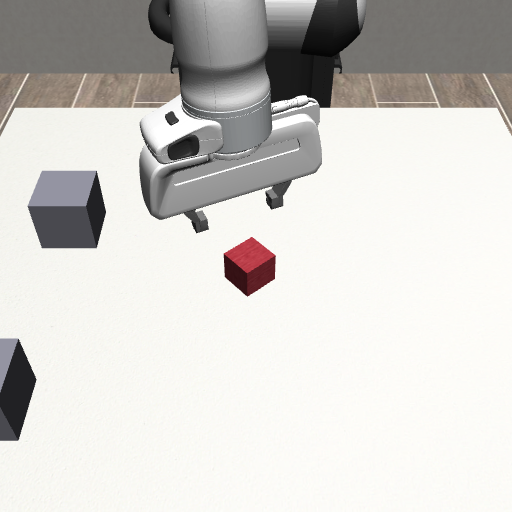} & \includegraphics[width=0.155\linewidth]{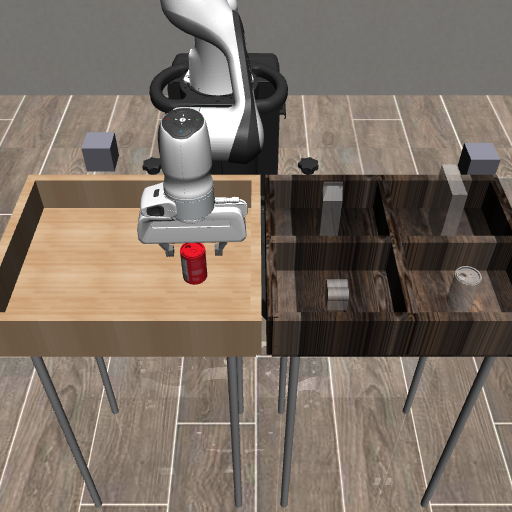} & \includegraphics[width=0.155\linewidth]{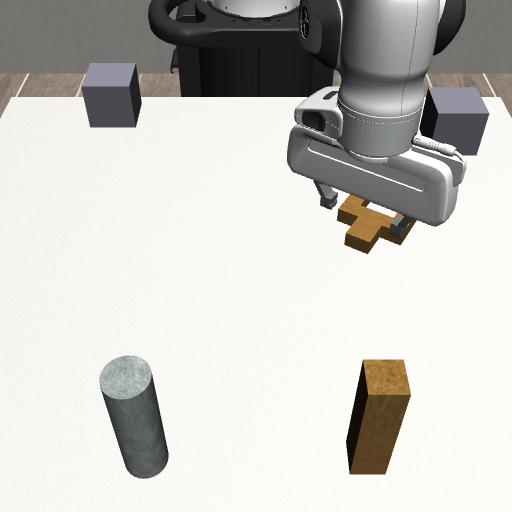} & \includegraphics[width=0.155\linewidth]{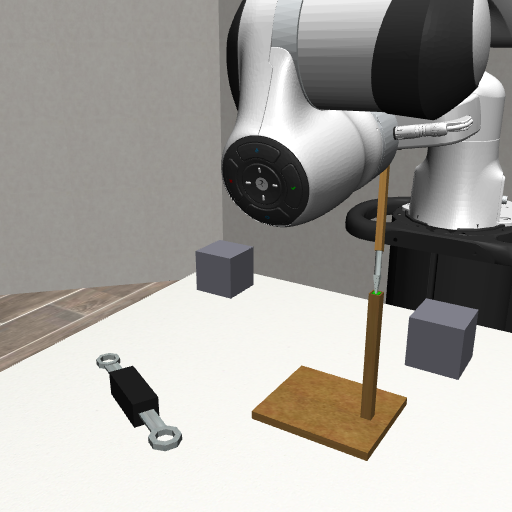} & \includegraphics[width=0.155\linewidth]{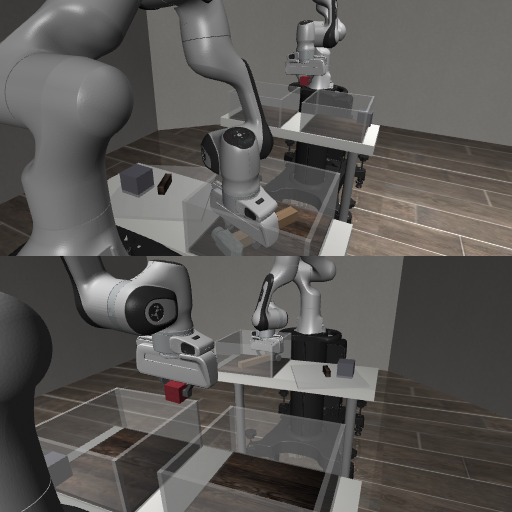} \\
PushT & Lift & Can & Square & ToolHang & Transport \\
\end{tabular}
\caption{The six tasks as the models see them, with their distractors: painted dot and bar on PushT, two wobbling boxes on the robomimic tasks (cameras and resolutions in \Cref{app:rm-config}).}
% \vspace{-4mm}
\label{fig:tasks}
\end{figure}

% \vspace{-2mm}
\section{Experiments and Results}
\label{sec:expts}
% \vspace{-2mm}

\subsection{Tasks and data.}
\label{subsec:tasks}
\emph{PushT}~\citep{florence2022implicit,chi2023diffusion}: a circular agent pushes a T-shaped block to a fixed target pose; actions are absolute target positions at 10\,Hz in chunks of five. We use the 206 human demonstrations of \citet{chi2023diffusion} and paint two distractors that are independent of the actions: a dot on a Lissajous curve and a bar sliding along the top edge.
\emph{robomimic}~\citep{mandlekar2021robomimic}: Lift, Can, Square, ToolHang and Transport, using the benchmark's 200 proficient-human demonstrations replayed in robosuite. We add two grey distractor boxes, larger than the task objects, that wobble in every scene without affecting the dynamics.

Every model is trained on these frames alone (\Cref{fig:tasks}): pixels in, actions out, with no proprioception or state. Training runs for 300 epochs and five seeds on a 90\% split of the data; evaluation uses the remaining 10\% and runs in the same simulator with distractors present.

\subsection{Baselines.} 
\label{subsec:baselines}
We train three baselines and a reference. All are our reimplementations that retain their authors' original architectures but share one training protocol (pixels only, 300 epochs, five seeds, identical optimization and learning-rate schedules), sized to roughly match \lewam{}'s 41.0M trainable parameters. \emph{Dreamer}~\citep{hafner2020dreamer}: an RSSM trained by image reconstruction with an MLP action head. \emph{LeWM}~\citep{lewm2026}: has no action head, so we freeze it and train \lewam{}'s rectified-flow head on its latents with the same loss and schedule as \lewam{}. \emph{\genwam{}}: the Cosmos Policy architecture~\citep{nvidia2025cosmospolicy} at \lewam{}'s scale, a flow-matching transformer over the tokens of a VAE pretrained on each dataset's frames, denoising the next frame and action chunk jointly. \emph{Flow-matching policy}: \lewam{}'s encoder and head trained through the policy loss only, so no world model.

\subsection{Latent Space analysis}
\label{subsec:latent_analysis}
For each model and task we fit a ridge regression from the latent to agent, object, gripper and distractor state, with cross-validation to remove any decoder confounds, and evaluate it on a hold-out set. We track MSE, Pearson's $r$ and explained variance $R^2$, both on the state variables that matter for the task and on those of the task-irrelevant distractors. \Cref{tab:probe_all_metrics_ep300} shows that, on average, \lewam{} encodes task-relevant state better than LeWM and even the reconstruction-based models, and ignores distractors as well as LeWM.

% \begin{table}[t]\centering\small\setlength{\tabcolsep}{4pt}
% % \vspace{-3mm}

% \caption{Linear probe metrics across all tasks, 5 seeds (Negative $R^2$ clipped to 0).
% }
% \label{tab:probe_all_metrics_ep300_avg}
% \begin{tabular}{lcccccc}\toprule
%  & \multicolumn{2}{c}{$R^2 = 1 - \mathrm{MSE}/\mathrm{Var}$} & \multicolumn{2}{c}{Pearson $r$} & \multicolumn{2}{c}{MSE} \\ \cmidrule(lr){2-3}\cmidrule(lr){4-5}\cmidrule(lr){6-7}
% \textbf{Model} & Task-state $\uparrow$ & Distractors $\downarrow$ & Task-state $\uparrow$ & Distractors $\downarrow$ & Task-state $\downarrow$ & Distractors $\uparrow$ \\ \midrule
% % Policy-only FM & \textbf{0.80 $\pm$ 0.01} & 0.08 $\pm$ 0.01 & 0.88 $\pm$ 0.01 & 0.28 $\pm$ 0.01 & \textbf{0.21} $\pm$ 0.01 & 0.92 $\pm$ 0.01 \\
% Dreamer & 0.68 $\pm$ 0.00 & 0.19 $\pm$ 0.01 & 0.83 $\pm$ 0.00 & 0.34 $\pm$ 0.01 & 0.32 $\pm$ 0.00 & 0.81 $\pm$ 0.01 \\
% LeWM + diff. head & 0.62 $\pm$ 0.01 & \textbf{0.00 $\pm$ 0.01} & 0.78 $\pm$ 0.01 & \textbf{0.03 $\pm$ 0.01} & 0.39 $\pm$ 0.01 & 1.04 $\pm$ 0.01 \\
% RecWAM & 0.78 $\pm$ 0.01 & 0.73 $\pm$ 0.01 & 0.87 $\pm$ 0.00 & 0.85 $\pm$ 0.00 & 0.22 $\pm$ 0.00 & 0.27 $\pm$ 0.01 \\
% \textbf{LeWAM (ours)} & \textbf{0.80 $\pm$ 0.00} & \textbf{0.00 $\pm$ 0.01} & \textbf{0.89 $\pm$ 0.00} & \textbf{0.03 $\pm$ 0.01} & \textbf{0.21 $\pm$ 0.00} & \textbf{1.10 $\pm$ 0.01} \\
% \bottomrule
% % \vspace{-2mm}
% \end{tabular}
% \end{table}

\begin{table}[t]\centering\small\setlength{\tabcolsep}{4pt}
% \vspace{-3mm}

% \caption{Linear probe metrics per task, 5 seeds (Negative $R^2$ clipped to 0).}
\caption{Linear probe metrics across all tasks, 5 seeds (Negative $R^2$ clipped to 0).}

\label{tab:probe_all_metrics_ep300}
\begin{tabular}{lcccccc}\toprule
 & \multicolumn{2}{c}{$R^2 = 1 - \mathrm{MSE}/\mathrm{Var}$} & \multicolumn{2}{c}{Pearson $r$} & \multicolumn{2}{c}{MSE} \\ \cmidrule(lr){2-3}\cmidrule(lr){4-5}\cmidrule(lr){6-7}
\textbf{Model} & Task-state $\uparrow$ & Distractors $\downarrow$ & Task-state $\uparrow$ & Distractors $\downarrow$ & Task-state $\downarrow$ & Distractors $\uparrow$ \\ \midrule
\multicolumn{7}{l}{\textit{PushT}} \\
% Policy-only FM & \textbf{0.85 $\pm$ 0.01} & 0.01 $\pm$ 0.04 & \textbf{0.93 $\pm$ 0.00} & 0.27 $\pm$ 0.04 & \textbf{0.13 $\pm$ 0.00} & 1.00 $\pm$ 0.04 \\
Dreamer & 0.56 $\pm$ 0.01 & 0.03 $\pm$ 0.03 & 0.78 $\pm$ 0.00 & 0.30 $\pm$ 0.01 & 0.39 $\pm$ 0.01 & 0.96 $\pm$ 0.03 \\
LeWM + diff. head & 0.54 $\pm$ 0.05 & \textbf{0.00 $\pm$ 0.02} & 0.77 $\pm$ 0.03 & \textbf{0.06 $\pm$ 0.04} & 0.40 $\pm$ 0.05 & 1.07 $\pm$ 0.02 \\
RecWAM & 0.64 $\pm$ 0.02 & 0.79 $\pm$ 0.03 & 0.82 $\pm$ 0.02 & 0.89 $\pm$ 0.02 & 0.32 $\pm$ 0.02 & 0.21 $\pm$ 0.03 \\
\textbf{LeWAM (ours)} & \textbf{0.81 $\pm$ 0.01} & \textbf{0.00 $\pm$ 0.02} & \textbf{0.92 $\pm$ 0.00} & \textbf{0.06 $\pm$ 0.02} & \textbf{0.16 $\pm$ 0.01} & \textbf{1.19 $\pm$ 0.02} \\
\midrule
\multicolumn{7}{l}{\textit{Lift}} \\
% Policy-only FM & 0.41 $\pm$ 0.04 & \textbf{0.00 $\pm$ 0.05} & 0.64 $\pm$ 0.03 & 0.28 $\pm$ 0.02 & 0.65 $\pm$ 0.04 & 1.02 $\pm$ 0.05 \\
Dreamer & 0.46 $\pm$ 0.01 & 0.81 $\pm$ 0.02 & 0.68 $\pm$ 0.00 & 0.90 $\pm$ 0.01 & 0.60 $\pm$ 0.01 & 0.19 $\pm$ 0.02 \\
LeWM + diff. head & 0.26 $\pm$ 0.02 & \textbf{0.00 $\pm$ 0.02} & 0.52 $\pm$ 0.02 & \textbf{0.01 $\pm$ 0.03} & 0.81 $\pm$ 0.02 & 1.08 $\pm$ 0.02 \\
RecWAM & \textbf{0.56 $\pm$ 0.01} & 0.86 $\pm$ 0.01 & \textbf{0.72 $\pm$ 0.01} & 0.93 $\pm$ 0.00 & \textbf{0.48 $\pm$ 0.01} & 0.14 $\pm$ 0.01 \\
\textbf{LeWAM (ours)} & 0.42 $\pm$ 0.02 & \textbf{0.00 $\pm$ 0.05} & 0.67 $\pm$ 0.01 & 0.16 $\pm$ 0.04 & 0.63 $\pm$ 0.02 & \textbf{1.20 $\pm$ 0.05} \\
\midrule
\multicolumn{7}{l}{\textit{Can}} \\
% Policy-only FM & 0.82 $\pm$ 0.01 & 0.42 $\pm$ 0.04 & 0.91 $\pm$ 0.01 & 0.66 $\pm$ 0.03 & 0.18 $\pm$ 0.01 & 0.59 $\pm$ 0.04 \\
Dreamer & 0.79 $\pm$ 0.01 & 0.01 $\pm$ 0.03 & 0.89 $\pm$ 0.01 & 0.17 $\pm$ 0.05 & 0.21 $\pm$ 0.01 & 0.99 $\pm$ 0.03 \\
LeWM + diff. head & 0.64 $\pm$ 0.01 & \textbf{0.00 $\pm$ 0.02} & 0.81 $\pm$ 0.00 & \textbf{0.02 $\pm$ 0.02} & 0.37 $\pm$ 0.01 & 1.04 $\pm$ 0.02 \\
RecWAM & \textbf{0.84 $\pm$ 0.01} & 0.61 $\pm$ 0.02 & \textbf{0.92 $\pm$ 0.01} & 0.78 $\pm$ 0.01 & \textbf{0.16 $\pm$ 0.01} & 0.39 $\pm$ 0.02 \\
\textbf{LeWAM (ours)} & \textbf{0.84 $\pm$ 0.00} & \textbf{0.00 $\pm$ 0.02} & \textbf{0.92 $\pm$ 0.00} & 0.20 $\pm$ 0.03 & \textbf{0.17 $\pm$ 0.01} & \textbf{1.07 $\pm$ 0.02} \\
\midrule
\multicolumn{7}{l}{\textit{Square}} \\
% Policy-only FM & 0.89 $\pm$ 0.00 & 0.11 $\pm$ 0.04 & 0.94 $\pm$ 0.00 & 0.33 $\pm$ 0.03 & 0.11 $\pm$ 0.00 & 0.90 $\pm$ 0.04 \\
Dreamer & 0.70 $\pm$ 0.01 & 0.07 $\pm$ 0.02 & 0.84 $\pm$ 0.01 & 0.27 $\pm$ 0.03 & 0.28 $\pm$ 0.01 & 0.93 $\pm$ 0.02 \\
LeWM + diff. head & 0.72 $\pm$ 0.01 & \textbf{0.00 $\pm$ 0.00} & 0.85 $\pm$ 0.01 & \textbf{-0.01 $\pm$ 0.01} & 0.27 $\pm$ 0.01 & 1.02 $\pm$ 0.00 \\
RecWAM & 0.87 $\pm$ 0.01 & 0.61 $\pm$ 0.01 & 0.93 $\pm$ 0.00 & 0.77 $\pm$ 0.01 & 0.13 $\pm$ 0.01 & 0.40 $\pm$ 0.01 \\
\textbf{LeWAM (ours)} & \textbf{0.90 $\pm$ 0.01} & \textbf{0.00 $\pm$ 0.02} & \textbf{0.95 $\pm$ 0.00} & \textbf{0.03 $\pm$ 0.01} & \textbf{0.10 $\pm$ 0.01} & \textbf{1.08 $\pm$ 0.02} \\
\midrule
\multicolumn{7}{l}{\textit{ToolHang}} \\
% Policy-only FM & 0.92 $\pm$ 0.00 & \textbf{0.00 $\pm$ 0.01} & \textbf{0.96 $\pm$ 0.00} & 0.11 $\pm$ 0.04 & 0.08 $\pm$ 0.00 & 0.99 $\pm$ 0.01 \\
Dreamer & 0.76 $\pm$ 0.00 & 0.23 $\pm$ 0.03 & 0.87 $\pm$ 0.00 & 0.40 $\pm$ 0.00 & 0.24 $\pm$ 0.00 & 0.77 $\pm$ 0.03 \\
LeWM + diff. head & 0.82 $\pm$ 0.02 & \textbf{0.00 $\pm$ 0.00} & 0.91 $\pm$ 0.01 & \textbf{0.01 $\pm$ 0.01} & 0.18 $\pm$ 0.02 & \textbf{1.00 $\pm$ 0.00} \\
RecWAM & 0.88 $\pm$ 0.00 & 0.75 $\pm$ 0.01 & 0.94 $\pm$ 0.00 & 0.86 $\pm$ 0.01 & 0.12 $\pm$ 0.00 & 0.25 $\pm$ 0.01 \\
\textbf{LeWAM (ours)} & \textbf{0.93 $\pm$ 0.00} & \textbf{0.00 $\pm$ 0.00} & \textbf{0.96 $\pm$ 0.00} & \textbf{-0.02 $\pm$ 0.01} & \textbf{0.07 $\pm$ 0.00} & \textbf{1.01 $\pm$ 0.00} \\
\midrule
\multicolumn{7}{l}{\textit{Transport}} \\
% Policy-only FM & \textbf{0.90 $\pm$ 0.01} & \textbf{0.00 $\pm$ 0.00} & \textbf{0.95 $\pm$ 0.00} & 0.03 $\pm$ 0.03 & \textbf{0.10 $\pm$ 0.01} & \textbf{1.02 $\pm$ 0.00} \\
Dreamer & 0.80 $\pm$ 0.03 & \textbf{0.00 $\pm$ 0.00} & 0.89 $\pm$ 0.02 & 0.03 $\pm$ 0.02 & 0.21 $\pm$ 0.03 & 1.01 $\pm$ 0.00 \\
LeWM + diff. head & 0.72 $\pm$ 0.01 & \textbf{0.00 $\pm$ 0.00} & 0.84 $\pm$ 0.01 & \textbf{0.01 $\pm$ 0.01} & 0.29 $\pm$ 0.01 & 1.01 $\pm$ 0.00 \\
RecWAM & 0.86 $\pm$ 0.01 & 0.76 $\pm$ 0.01 & 0.92 $\pm$ 0.01 & 0.87 $\pm$ 0.01 & 0.15 $\pm$ 0.01 & 0.24 $\pm$ 0.01 \\
\textbf{LeWAM (ours)} & \textbf{0.88 $\pm$ 0.00} & \textbf{0.00 $\pm$ 0.00} & \textbf{0.94 $\pm$ 0.00} & \textbf{-0.02 $\pm$ 0.01} & \textbf{0.12 $\pm$ 0.00} & \textbf{1.02 $\pm$ 0.00} \\
\midrule\midrule
\multicolumn{7}{l}{\textit{Average over tasks}} \\
% Policy-only FM & \textbf{0.80 $\pm$ 0.01} & 0.08 $\pm$ 0.01 & 0.88 $\pm$ 0.01 & 0.28 $\pm$ 0.01 & \textbf{0.21} $\pm$ 0.01 & 0.92 $\pm$ 0.01 \\
Dreamer & 0.68 $\pm$ 0.00 & 0.19 $\pm$ 0.01 & 0.83 $\pm$ 0.00 & 0.34 $\pm$ 0.01 & 0.32 $\pm$ 0.00 & 0.81 $\pm$ 0.01 \\
LeWM + diff. head & 0.62 $\pm$ 0.01 & \textbf{0.00 $\pm$ 0.01} & 0.78 $\pm$ 0.01 & \textbf{0.03 $\pm$ 0.01} & 0.39 $\pm$ 0.01 & 1.04 $\pm$ 0.01 \\
RecWAM & 0.78 $\pm$ 0.01 & 0.73 $\pm$ 0.01 & 0.87 $\pm$ 0.00 & 0.85 $\pm$ 0.00 & 0.22 $\pm$ 0.00 & 0.27 $\pm$ 0.01 \\
\textbf{LeWAM (ours)} & \textbf{0.80 $\pm$ 0.00} & \textbf{0.00 $\pm$ 0.01} & \textbf{0.89 $\pm$ 0.00} & \textbf{0.03 $\pm$ 0.01} & \textbf{0.21 $\pm$ 0.00} & \textbf{1.10 $\pm$ 0.01} \\
\bottomrule
% \vspace{-8mm}
\end{tabular}
\end{table}

\begin{figure}[h]
\centering
% \vspace{-4mm}
\includegraphics[width=\linewidth]{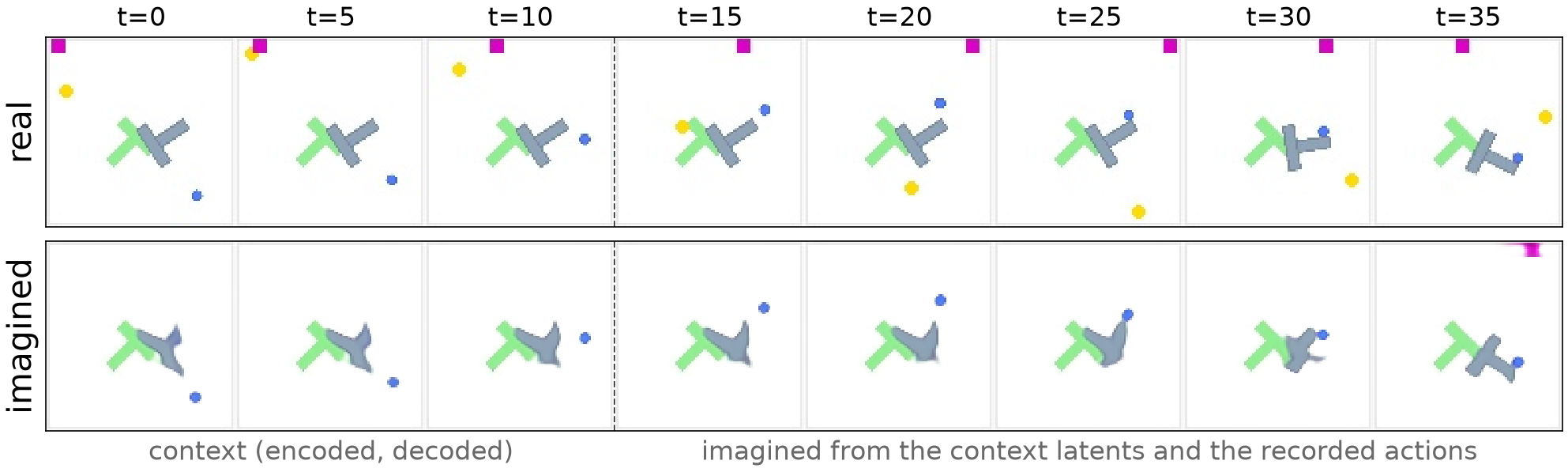}
\includegraphics[width=\linewidth]{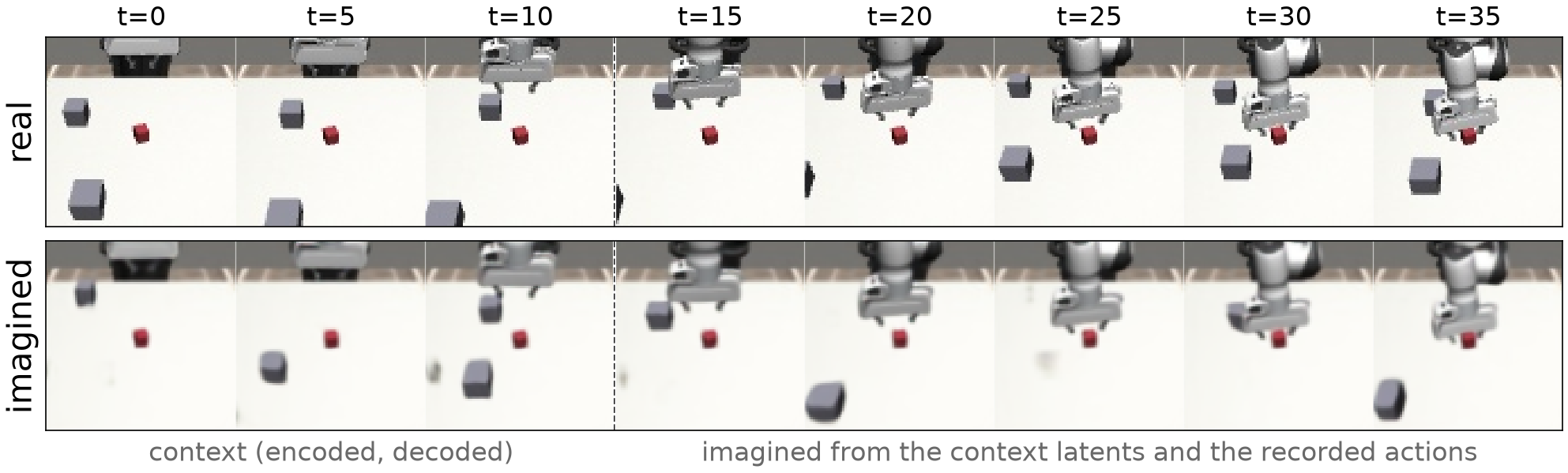}
% \vspace{-3mm}
\caption{Imagined rollouts with LeWAM using ground truth actions. The top image is for PushT and the bottom one is for the Lift Robomimic task. The first 3 frames are context frames that is passed as input to the model. The 5 frames that follow are imagined frames by rolling out the LeWAM's forward mode using ground truth actions.}
\label{fig:imagine}
\end{figure}

To visually verify the quality of the latent space and the world model, we train a plain feed-forward convolutional decoder on a frozen pre trained LeWAM model.
Note that gradients from the decoder do not change the latent space as this is done after LeWAM training.
We show the rollouts from PushT and Robomimic Lift in \Cref{fig:imagine} tasks.
For PushT we can see that the location of T block is consistently identified, but the rotation is blurry.
This seems to be true for both the input context frames as well as the imagined frames.
A point to note here is that the distractors are almost completely missing in the imagined rollouts.
This further strengthens the linear probe results.
In the Lift task, we see similar results, where the distractor blocks are either missing or wrongly predicted by the decoder.
We believe the presence of the distractor block (but still in the wrong location) is an artifact of the decoder hallucinating.
% \vspace{-6mm}

\subsection{Closed-loop evaluation.}
\label{subsec:cle}

We evaluate each model's policy for 250 rollouts (5 training $\times$ 50 evaluation seeds) on all six tasks with visual distractors. For each world model we also run MPC in its latent space, sampling both in action space (CEM, MPPI, CMA-ES) and in the policy's noise space (DS-CEM, DS-MPPI, DS-CMA-ES); the goal is the latent 25 steps after the demonstration frame nearest the current latent. A BC-only flow-matching policy serves as reference. 
\begin{wrapfigure}{r}{0.6\textwidth}
\centering
% \vspace{-6mm}
\includegraphics[width=\linewidth]{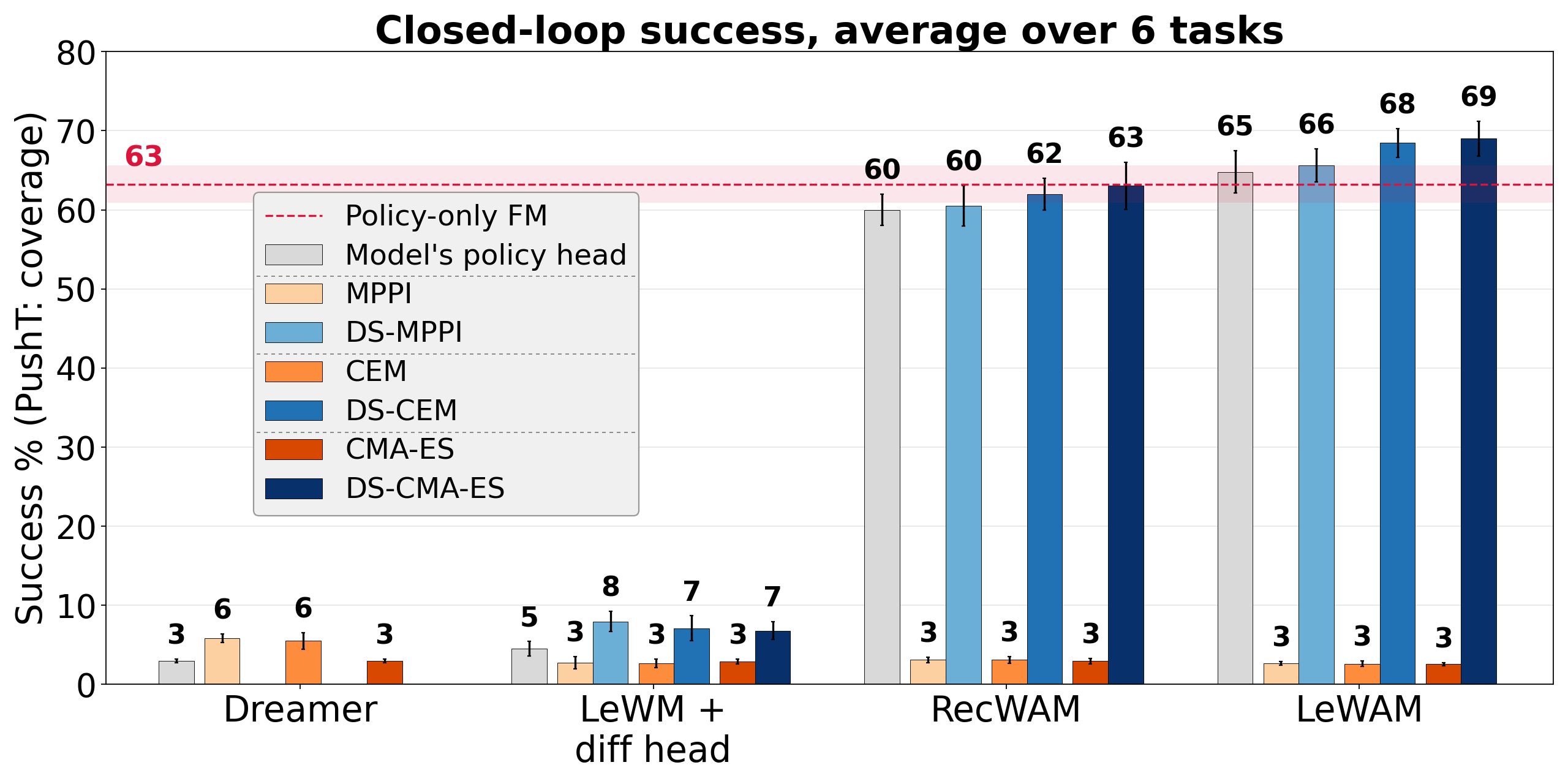}
% \vspace{-4mm}
\caption{Closed-loop success of all models and planners, over 5 training seeds $\times$ 50 evaluation seeds (250 rollouts).}
\label{fig:avg_cle}
% \vspace{-2mm}
\end{wrapfigure}

% \begin{figure}{t}
% \centering
% % \vspace{-4mm}
% \includegraphics[width=\linewidth]{figures/results/closed_loop_ep300_average.png}
% \vspace{-4mm}
% \caption{Closed-loop success of all models and planners, over 5 training seeds $\times$ 50 evaluation seeds (250 rollouts).}
% \label{fig:avg_cle}
% \vspace{-4mm}
% \end{figure}

\Cref{fig:avg_cle} shows that \genwam{}'s policy is slightly below the reference, which we believe reflects capacity spent on pixel reconstruction, while Dreamer and LeWM + diffusion head perform very poorly. \lewam{}'s policy maintains the reference's performance. Sampling in noise space gives both \lewam{} and \genwam{} a performance gain at inference time, whereas action-space MPC collapses every model to 3--8\% success, confirming that the planner exploits dynamics errors (\Cref{sec:break}).

\subsection{Ablation}
\label{subsec:ablation}

% \documentclass{article}
% \usepackage{booktabs}
% \usepackage{multirow}
% \usepackage[margin=1in]{geometry}
% \begin{document}

\begin{table}[h]
\centering
\caption{LeWAM mode ablation. 
One mode removed at a time from the four-mode model (41M, object-distractor datasets, 300 epochs). 
% Mean $\pm$ std over seeds 0--4. 
$R^2$: ridge probe from the frozen latent to task state, computed as $1 - \mathrm{MSE}/\mathrm{Var}$ on z-scored targets. 
Policy / DS-CEM: closed-loop success (\%), 50 episodes per cell. 
The no-policy variant has no policy head, so its DS-CEM column steers the inverse-trained flow head.}
\label{tab:mode_ablation}
\small
\begin{tabular}{llccc}
\toprule
Task & Variant & $R^2$ state $\uparrow$ & Policy $\uparrow$ & DS-CEM $\uparrow$ \\
\midrule
\multirow{5}{*}{Lift}
 & All four modes & $\mathbf{0.423 \pm 0.019}$ & $\mathbf{94.0 \pm 4.2}$ & $\mathbf{97.6 \pm 0.8}$ \\
 & No forward     & $0.323 \pm 0.045$ & $91.6 \pm 2.3$ & $95.6 \pm 2.9$ \\
 & No inverse     & $0.386 \pm 0.031$ & $92.0 \pm 5.5$ & $95.6 \pm 2.9$ \\
 & No policy      & $0.359 \pm 0.019$ & --             & $0.8 \pm 1.6$  \\
 & No backward    & $0.370 \pm 0.056$ & $88.0 \pm 4.6$ & $93.6 \pm 2.3$ \\
\midrule
\multirow{5}{*}{Can}
 & All four modes & $\mathbf{0.839 \pm 0.004}$ & $\mathbf{82.8 \pm 6.1}$ & $\mathbf{78.8 \pm 4.5}$ \\
 & No forward     & $0.781 \pm 0.018$ & $77.6 \pm 8.3$ & $74.4 \pm 7.7$ \\
 & No inverse     & $0.817 \pm 0.013$ & $79.6 \pm 8.0$ & $73.2 \pm 8.2$ \\
 & No policy      & $0.761 \pm 0.010$ & --             & $0.0 \pm 0.0$  \\
 & No backward    & $0.801 \pm 0.015$ & $77.6 \pm 3.9$ & $77.6 \pm 2.7$ \\
\midrule
\multirow{5}{*}{Square}
 & All four modes & $\mathbf{0.896 \pm 0.007}$ & $80.4 \pm 1.5$ & $\mathbf{87.2 \pm 4.7}$ \\
 & No forward     & $0.876 \pm 0.003$ & $71.2 \pm 8.4$ & $73.6 \pm 1.5$ \\
 & No inverse     & $0.885 \pm 0.003$ & $\mathbf{82.0 \pm 3.8}$ & $81.6 \pm 3.2$ \\
 & No policy      & $0.858 \pm 0.015$ & --             & $0.0 \pm 0.0$  \\
 & No backward    & $0.872 \pm 0.007$ & $74.0 \pm 7.7$ & $82.8 \pm 3.7$ \\
\bottomrule
\end{tabular}
\end{table}

To analyze the importance of each mode in LeWAM, we perform an ablation by switching off each mode during training and evaluate it on three robomimic tasks (Lift, Can and Square). 
From \Cref{tab:mode_ablation}, we can see that all modes add value to both forming an informative latent space as well as closed-loop performance on average over tasks.
State information in the latent space, policy rollout and planning all seem to improve with all four modes.
Forward and backward consistently help; inverse is neutral on Square.

\begin{table}[t]
  \centering
  % \caption{Projection ablation. DS-CEM closed-loop success (\%) with LeWAM,
  % with and without projecting proposals onto the $\sqrt{A}$ shell before
  % decoding (Section~\ref{sec:variants}). Same budget as Table~\ref{tab:mode_ablation}
  % ($64 \times 3$, horizon 5); mean $\pm$ std over seeds 0--4, 50 episodes per cell.
  % Without projection, the planner's mean drifts toward low-norm noise and the
  % decoded chunks become mode-seeking, which costs the most on Can.}
    \caption{Projection ablation. DS-CEM closed-loop success (\%) with \lewam{},
    with and without projecting proposals onto the $\sqrt{A}$ shell before
    decoding (\Cref{sec:variants}). 
    % Budget of \Cref{sec:ds} ($64 \times 3$, horizon 5);
    mean $\pm$ std over seeds 0--4, 50 episodes per cell. 
    % The projected column is the DS-CEM column of \Cref{tab:mode_ablation}.
    }  
  \label{tab:projection}
  \small
  \begin{tabular}{lcc}
    \toprule
    Task   & With $\sqrt{A}$ shell & Without shell \\
    \midrule
        Lift   & $\mathbf{97.6 \pm 0.8}$ & $93.0 \pm 3.0$ \\
        Can    & $\mathbf{78.8 \pm 4.5}$ & $23.0 \pm 8.0$ \\
        Square & $\mathbf{87.2 \pm 4.7}$ & $75.0 \pm 3.0$ \\
    \bottomrule
  \end{tabular}
\end{table}
We also ablate the projection step of \Cref{sec:variants}, keeping the budget and goals of \Cref{sec:ds}. Removing it lowers DS-CEM success on all three tasks (\Cref{tab:projection}), by about 5 points on Lift, 12 on Square and 56 on Can.

% \section{Imagined Rollouts}

% \section{Conclusion and Future work}

% This paper presents LeWAM, a decoder-free JEPA-based world-action model.
% We show that LeWAM's latent space is more informative than LeWM, and is void of the visual features that do not affect task dynamics.
% This property allows the LeWAM to better model policy actions.
% Finally, we also show that by sampling in diffusion noise space (an inference-only procedure), we can effectively plan with LeWAM and improve average closed-loop performance across tasks.
% We believe that the two logical next steps are: 1. applying LeWAM on real-world data (that consists of an abundance of dynamically inert features); 2. focus on improving planning algorithms with LeWAM, as these are slow and also require a notion of a goal at inference time.

\section{Conclusion and Future work}

This paper presents \lewam{}, a decoder-free JEPA-based world action model.
We show that \lewam{}'s latent space is more informative than LeWM's, and is void of the visual features that do not affect task dynamics.
This property allows \lewam{} to better model policy actions.
Finally, we show that by sampling in diffusion noise space (an inference-only procedure), we can effectively plan with \lewam{} and improve average closed-loop performance across tasks.
We believe the two logical next steps are: (1) applying \lewam{} to real-world data, which contains an abundance of dynamically inert features; and (2) improving planning algorithms with \lewam{}, as these are slow and also require a notion of a goal at inference time.
%===============================================================================

% %===============================================================================

\clearpage

%===============================================================================

\bibliography{example}  % .bib
\appendix

\section{Results per Task}
\label{app:probe_per_task}

\begin{figure}[h]
\centering
\includegraphics[width=\linewidth]{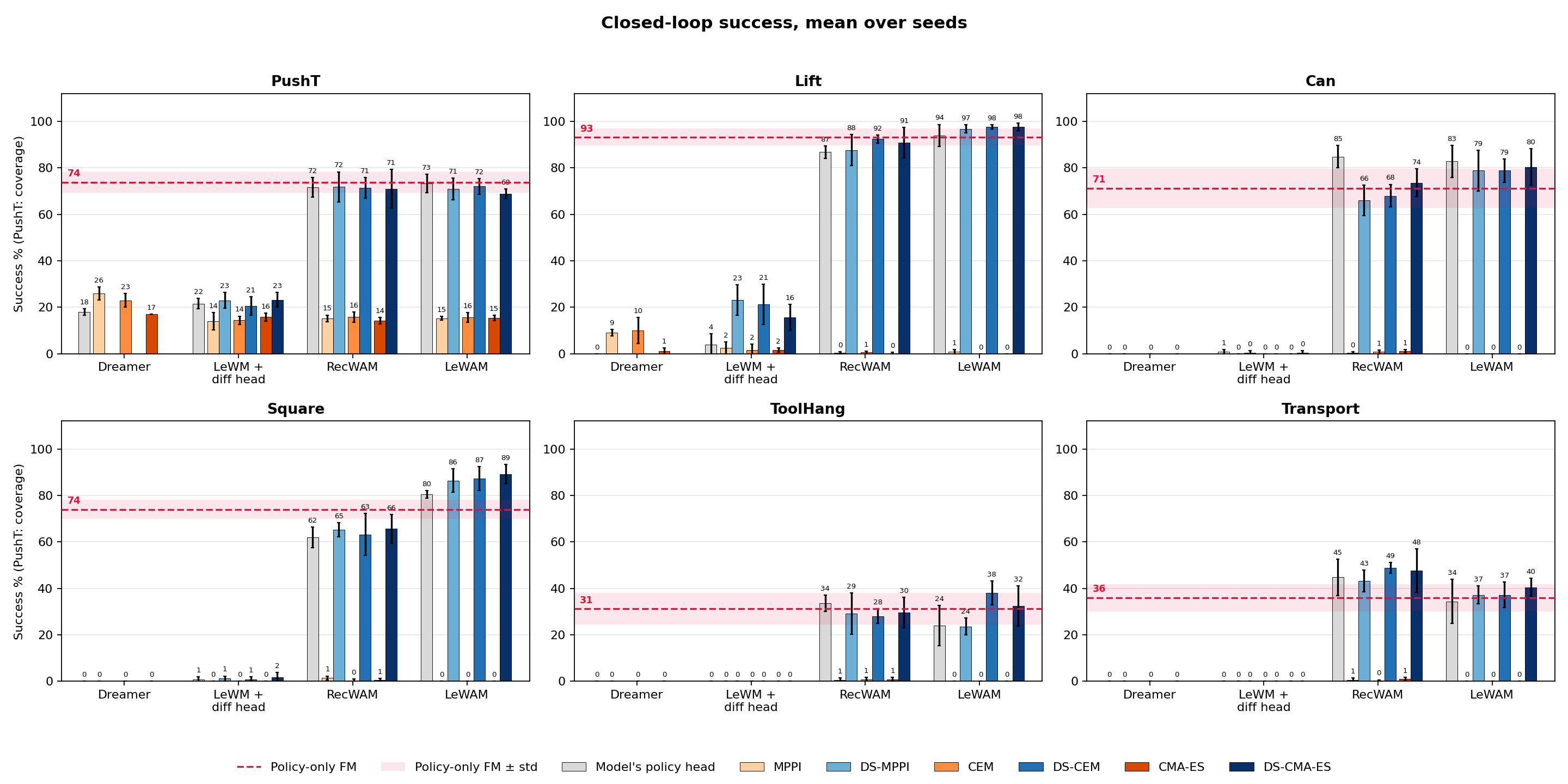}
\caption{Average closed-loop performance of all models/MPC, for each task, over 5 training seeds, 50 evaluation seeds: 250 unique rollouts.}
\label{fig:cle_per_task}
\end{figure}

% While \Cref{tab:probe_all_metrics_ep300_avg} shows results averaged across all tasks, we show results per task in \Cref{tab:probe_all_metrics_ep300_per_task} for a more comprehensive look. 
While \Cref{fig:avg_cle} shows closed loop evaluation results averaged across all tasks, we show the per-task closed loop evaluation and MPC results in \Cref{fig:cle_per_task}.

% \begin{sidewaystable}[p]
% \caption{}
% \label{tab:probes}
% \centering\small\setlength{\tabcolsep}{4pt}
% \input{tables/300_combined_probe_task}
% \end{sidewaystable}

% \begin{figure}[h]
% \centering
% \includegraphics[width=\linewidth]{Corl_8_page_workshop/figures/results/can_imagination_holdout.png}
% \includegraphics[width=\linewidth]{Corl_8_page_workshop/figures/results/square_imagination_holdout.png}
% \includegraphics[width=\linewidth]{Corl_8_page_workshop/figures/results/tool_hang_imagination_holdout.png}
% \includegraphics[width=\linewidth]{Corl_8_page_workshop/figures/results/transport_imagination_holdout.png}
% \caption{Imagined rollouts for tasks not shown in the main paper.}
% \label{fig:imagine_per_task}
% \end{figure}

% \input{sections/appendix_sections/per_task_cle}

\section{Configuration of the tasks}
\label{app:rm-config}
\paragraph{Cameras and resolutions.} PushT is rendered at $96$ px on a $512$ px table and resized to $224$ px for the encoder. Lift, Can and Square are rendered at $84$ px from the agent-view camera; ToolHang at $168$ px from the benchmark's side camera, which keeps the tool, its frame and the boxes in view; and Transport at $256$ px as the two shoulder cameras, one per arm, stacked into a single frame so that the hand-over between the arms is visible. All frames are resized once to $224$ px. Evaluation uses 50 seeded simulator resets per seed, with the boxes moving.

\paragraph{Configuration.} \Cref{tab:rm-config} is the single \lewam{} configuration used on all tasks. 
Note: the training window's context is truncated to a random length (one to three frames, left-aligned, the newest frame always present) instead of masked, so that the first decision of an episode, where the simulator repeats the reset frame, is supervised.
\begin{table}[H]
\caption{\lewam{} on the robomimic tasks: one configuration for Lift, Can, Square, ToolHang and Transport.}
\label{tab:rm-config}
\centering\small
\begin{tabular}{@{}l p{0.8\textwidth}@{}}
\toprule
data & 200 demonstrations per task, frames resized once to 224 px, frameskip 5, windows of 3 context frames + 1 target, 5-step action chunks (35-D; 70-D on Transport), z-scored \\
schedule & 300 epochs, batch 128, AdamW (lr $5 \times 10^{-5}$, weight decay $10^{-3}$), linear warm-up then cosine, bf16, gradient clip 1.0 \\
encoder & ViT-tiny/14 at 224 px, [CLS] token, projector MLP (hidden 2048) to a 384-D latent \\
regularizer & SIGReg on the frame latent, weight 0.36 (0.09 $\times$ the four modes), 17 knots, 1024 projections \\
backbone & bidirectional transformer, depth 8, width 512, 16 heads $\times$ 64, MLP 2048 \\
modes & all four on every batch; forward mode weight 2, inverse, policy and backward weight 1 \\
context & random context length (1 to 3 frames, left-aligned truncation); no context-frame masking \\
policy head & flow matching, 8 Euler steps at inference \\
evaluation & 50 seeded starts per seed, the benchmark horizon per task, boxes moving \\
% planners as in \Cref{tab:hparams} with a five-chunk horizon and the retrieved goal 25 steps ahead \\
\bottomrule
\end{tabular}
\end{table}

\section{Model sizes}
\label{app:compute}
All four models are matched in trainable parameters. \lewam{} has 41.0M: a ViT-tiny/14 encoder (width 192, depth 12, 3 heads $\times$ 64) with its projector, and a bidirectional transformer of width 512, depth 8, 16 heads $\times$ 64 and MLP 2048. LeWM's autoregressive predictor is widened from the released configuration (width 192, depth 6; 18.0M in all) to width 512, 16 heads $\times$ 64, MLP 2048 at depth 6, for 42.2M in all. \genwam{}'s flow transformer is widened and deepened to width 512, depth 10, 16 heads $\times$ 64, MLP 2048, for 42.6M trainable parameters, plus a 3.9M VAE tokenizer that is pretrained per dataset and frozen. Dreamer's RSSM is trained by image reconstruction at 96 px with the balanced KL, and is enlarged from the standard configuration (13.6M) by widening every component: deterministic state 512 to 1536, stochastic state 32 to 64, GRU and MLP hidden 512 to 1280, image embedding 1024 to 1280, for 43.4M.

\section{Protocol and training hyperparameters}
\label{app:hparams}
\paragraph{Probes.} Ridge regression from the latent to agent location, block location, block angle and the two distractor positions on PushT, and to the end-effector position, the gripper opening, the task object's position and the two boxes on robomimic; $R^2 = 1 - \mathrm{MSE}/\mathrm{Var}$ on held-out episodes, with five-fold cross-validation over the penalty and a shuffled-target control, the same regression fit to permuted targets, which sits at or below zero for every model and task.

\paragraph{Planning protocol details.} Every planner uses the same budget, 64 samples, 3 iterations and 8 elites: the action-space planners, CEM, MPPI and CMA-ES, over action sequences of 5 chunks.
% \dscem{}, DS-MPPI and DS-CMA-ES over the projected noise with a horizon of 2 chunks; 
% best-of-64 ranking draws 64 chunks from the policy head and executes the one with lowest $\hat c$. 
For Dreamer, whose MLP head is deterministic, the 64 ranked candidates are the head's chunk plus 63 Gaussian perturbations of it with standard deviation 0.2 in z-scored action units, scored by Dreamer's own planning cost, the pixel error of the decoded rollout to the goal frame; the head's chunk is always a candidate, so ranking cannot fall below the policy, and no \ds{} variant exists for it, since there is no noise-to-action map to search over. 
Goals are retrieved in each model's own latent space, by encoding the observation 25 steps ahead of the closest latent from a given demonstration. 
% Because a latent that carries the distractors could in principle retrieve neighbours matched on box position rather than on task state, we also ran the retrieval in the simulator's ground-truth state, a space shared by all models; the closed-loop results agree with the latent-space retrieval within one seed spread, for \genwam{} on Can as for every other model and task, so we report the latent-space retrieval, which is what a deployed system has, and \genwam{}'s collapse on Can under \ds{} (Section~\ref{sec:paper-protocol}) is not a retrieval artefact. 
The robomimic horizons are 100 steps on Lift, 400 on Can and Square, and 700 on ToolHang and Transport; \lewam{} uses the configuration of \Cref{tab:rm-config} on every robomimic task.

\end{document}